\documentclass[letter, 10pt, conference]{ieeeconf}      

\IEEEoverridecommandlockouts                              
\usepackage[utf8]{inputenc}
\usepackage[T1]{fontenc}
\usepackage{times,multirow,float}
\usepackage{graphicx}
\usepackage{epsfig,xspace,layout}
\usepackage{color}
\usepackage{amsfonts}
\usepackage{times}
\usepackage{amssymb}
\usepackage{amsmath}
\usepackage{hyperref}
\usepackage{amsmath,bm}
\usepackage{rotating}
\usepackage{mathrsfs}
\usepackage{mathtools}
\usepackage{makeidx}
\usepackage[]{threeparttable}
\usepackage{dsfont}
\usepackage{cite}
\usepackage{xcolor}
\usepackage{cleveref}
\usepackage{algorithm}
\usepackage{algpseudocode}
\usepackage{tikz}
\usetikzlibrary{shapes,arrows,positioning,calc}
\usepackage{siunitx}
\usepackage{multicol,lipsum}
\usepackage{subcaption}
\usepackage[font=footnotesize]{caption}
\usepackage{url}
\usepackage{booktabs}
\usepackage{lineno}
\usepackage{hhline}
\usepackage[Symbol]{upgreek}
\usepackage{float}


\title{\LARGE \bf
Learning Agile Flight Maneuvers: Deep SE(3) Motion Planning and Control for Quadrotors
}

\author{Yixiao Wang$^{*1}$, Bingheng Wang$^{*1}$, Shenning Zhang$^{1}$, Han Wei Sia$^{2}$, and Lin Zhao$^{1}$
\thanks{
\setlength{\fboxrule}{0.4pt}\setlength{\fboxsep}{6pt}%
  \fbox{\parbox{0.92\columnwidth}{
    Accepted for publication in the Proceedings of the 2023 IEEE
    International Conference on Robotics and Automation (ICRA).
    \copyright~2023 IEEE. Personal use of this material is permitted.
    Permission from IEEE must be obtained for all other uses, in any
    current or future media, including reprinting/republishing this
    material for advertising or promotional purposes, creating new
    collective works, for resale or redistribution to servers or
    lists, or reuse of any copyrighted component of this work in
    other works. DOI: 10.1109/ICRA48891.2023.10160712}}
$\ast$ The first two authors contributed equally.}
\thanks{This work was supported by the Singapore Ministry of Education AcRF Tier 1 (A-0009030-01-00). (\textit{Corresponding author: Lin Zhao}.)} 
\thanks{$1$ These authors are with the Department of Electrical and Computer Engineering, National University of Singapore, 4 Engineering Drive 3, 117583 Singapore, Singapore
        {\tt\small $\left\{ {} \right.$wangyixiao, wangbingheng, shenningzhang$\left.\right\}$@u.nus.edu}, {\tt\small elezhli@nus.edu.sg}}
\thanks{$2$ Han Wei Sia is with ST Engineering, 1 Ang Mo Kio Electronics Park Road, 567710 Singapore, Singapore
    {\tt\small siahanwei@hotmail.com}}%
}

\begin{document}

\maketitle
\thispagestyle{empty}
\pagestyle{empty}

\begin{abstract}
Agile flights of autonomous quadrotors in cluttered environments require constrained motion planning and control subject to translational and rotational dynamics. Traditional model-based methods typically demand complicated design and heavy computation. In this paper, we develop a novel deep reinforcement learning-based method that tackles the challenging task of flying through a dynamic narrow gate. We design a model predictive controller with its adaptive tracking references parameterized by a deep neural network (DNN). These references include the traversal time and the quadrotor SE(3) traversal pose that encourage the robot to fly through the gate with maximum safety margins from various initial conditions. To cope with the difficulty of training in highly dynamic environments, we develop a reinforce-imitate learning framework to train the DNN efficiently that generalizes well to diverse settings. Furthermore, we propose a binary search algorithm that allows online adaption of the SE(3) references to dynamic gates in real-time. Finally, through extensive high-fidelity simulations, we show that our approach is adaptive to different gate trajectories, velocities, and orientations.  
\end{abstract}

\section{Introduction}
Agile flights of autonomous quadrotors in cluttered environments require constrained motion planning and control in the SE(3) space (special euclidean group)~\cite{lee2010geometric, mellinger2011minimum, mueller2015computationally,pereira2021nonlinear, han2021fast}. It is challenging due to their underactuated nature which couples the translational and rotational dynamics. It is further complicated by the dynamic environments that impose complex constraints and demand the adaptation of trajectories in real-time. 

As a representative task, flying through narrow gates with quadrotors has attracted increasing research interest. Earlier works split the traversing process into several phases and proposed different methods to plan a sequence of trajectory segments with given constraints on the quadrotor attitude~\cite{mellinger2012trajectory,loianno2016estimation,falanga2017aggressive}. These constraints are hand-picked from extensive experimental trials, which only work for specific static scenarios. By considering the attitude constraints explicitly, Liu et al.~\cite{liu2018search} proposed a search-based trajectory planning method on SE(3). It models the quadrotor as an ellipsoid and searches for a collision-free trajectory from a set of motion primitives. However, the search-based approach generally suffers from high computational complexity. Direct nonlinear optimization with soft SE(3) constraints was adopted in \cite{wang2022geometrically}, which relies on various reparameterizations and transforms for fast computation and still only considers static gates.
\begin{figure}[t!]
	\centering
	{\includegraphics[width=0.4\textwidth]{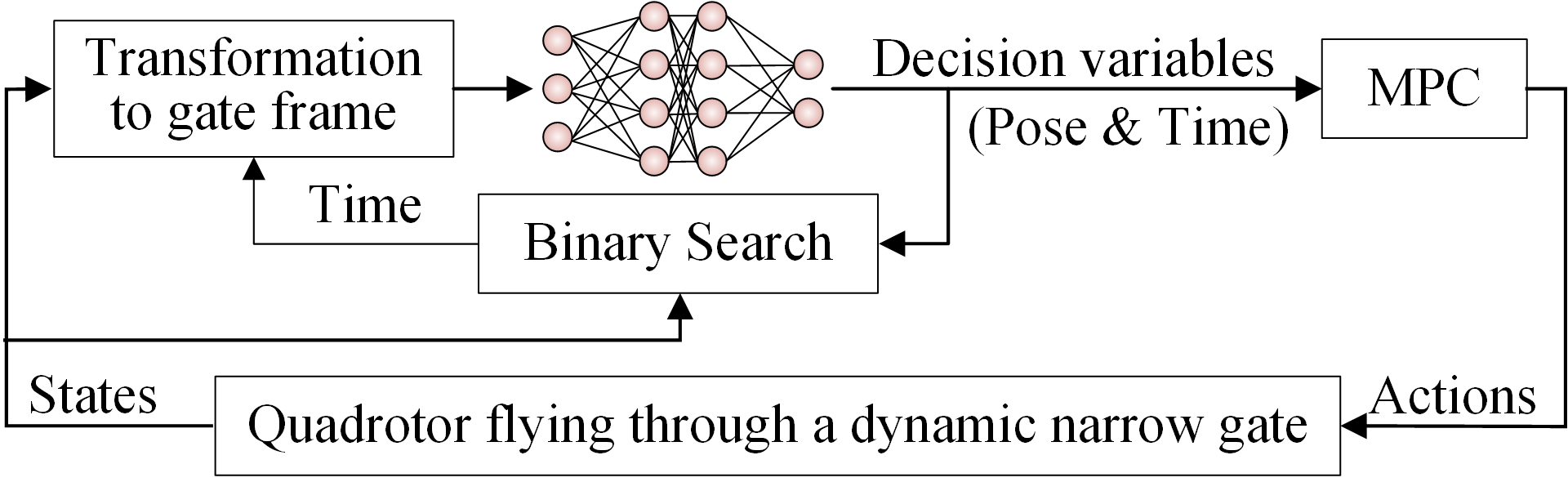}}
	\caption{\footnotesize Diagram of the proposed deep SE(3) motion planning and control method, which fuses a DNN with an MPC to address the gate traversing problem. The DNN adapts the SE(3) decision variables online for the MPC. Moreover, the binary search algorithm updates the DNN-predicted traversal time in real-time and is particularly effective for tracking a dynamic gate.}
\label{fig:overview}	
\end{figure}

All the above methods adopt the traditional two-layered framework that addresses planning and control separately. This framework generally works well for regular flights but can lead to inconsistencies between layers for agile flights, thus causing performance degradation in highly dynamic environments~\cite{shim2003decentralized}. In comparison, reinforcement learning (RL)-based methods have shown great potential in achieving agile flights as they can seamlessly integrate planning and control without excessive priors on desired trajectories. The authors in~\cite{lin2019flying} and \cite{xiao2021flying} showed that end-to-end deep neural networks (DNNs) enable a quadrotor to cross a static gate. While effective for a specific range of gate poses, these model-free RL methods generally lack the robustness that model-based methods have against uncertainties. Song et al.~\cite{song2022policy} proposed a hybrid method that trains a DNN to predict the traversal time as a decision variable for a model predictive controller (MPC). This method demonstrated efficacy in flying through a dynamic gate.  However, it fixes the quadrotor's attitude to be always aligned with the gate orientation, which is conservative and limits the agility. 

Driven by the dynamic narrow gate-traversing task, this paper proposes a novel deep SE(3) motion planning and control method for quadrotors. Specifically, we learn an MPC's SE(3) tracking references (meta-learning decision variables). These variables are parameterized by a portable DNN, including the traversal time and the quadrotor traversal pose, which can adapt online to diverse quadrotor states and gate parameters. The DNN is trained such that the generated references encourage the quadrotor to fly through a random gate with maximum safety margins, which fully exploits its agility. However, the receding horizon manner of MPC makes this learning problem computationally expensive; the dynamic environments can cause even unstable learning. To address these difficulties, we first learn the decision variables to traverse through a static gate and develop a reinforce-imitate learning framework that divides the training process into two tractable subproblems. Technical-wise, we train a first DNN via RL to generate the decision variables for a single-run open-loop MPC from stationary initial states. We then train a second DNN via supervised learning to imitate the first while making it adaptive to non-static states on the optimal MPC trajectory generated using the first DNN. Thus, the trained second DNN can generate the adaptive decision variables from dynamic states for an MPC in the receding horizon manner. Finally, we apply the MPC and the trained second DNN to a dynamic gate. As outlined in Fig.\ref{fig:overview}, we view the gate as "static" at its traversal position and transform the network inputs into the gate frame of that position. We develop a binary search algorithm to predict the gate traversal position by updating the traversal time from the second DNN. Given the current state feedback, it can update the predictions in real-time. This property is particularly effective for improving the adaptation of the MPC to dynamic environments.

Our main contributions are summarized as follows:
\begin{enumerate}
    \item We propose a novel deep SE(3) motion planning and control method for quadrotors and demonstrate its effectiveness via challenging tasks of traversing through fast-moving and rotating narrow gates in real-time under various settings;
    \item We develop a computationally efficient reinforce-imitate learning framework for training the DNN that parameterizes the traversal SE(3) decision variables, which generalizes well to unseen environments with different initial quadrotor states, gate widths and orientations, etc;
    \item We propose a binary search algorithm that enables fast real-time adaptation to dynamic gates of different moving and angular velocities;
    \item We further validate the robustness of the proposed methods in high-fidelity simulations that consider more complex aerodynamics than training. 
\end{enumerate}

The rest of this paper is organized as follows. Section~\ref{sec:problem} formulates an MPC for the gate-traversing problem. Section~\ref{sec:learning framework} develops the reinforce-imitate learning framework. The binary search algorithm is designed in Section~\ref{sec:binary search}. Simulation results are demonstrated in Section~\ref{sec:experiment}. We conclude this paper and discuss our future work in Section~\ref{sec:conclusion}.

\section{Model Predictive Control for Planning Agile Flight Maneuvers on SE(3)}\label{sec:problem}

\subsection{Quadrotor Model}\label{subsec:quadrotor}
We model the quadrotor as a 6 degree-of-freedom (DoF) rigid body of mass $m$ and moment of inertia $\mathbf{J}\in \mathbb{R}^{3\times 3}$. The equations governing the system motions can be written as
\begin{subequations}
\begin{align}
\dot{\mathbf{p}}_w &=\mathbf{v}_w, & \dot{\mathbf{v}}_w&=m^{-1}\mathbf{R}\left ( \mathbf{q} \right )f_\mathrm{r}\mathbf{e}_z-g\mathbf{e}_z
\label{eq:position dynamics},\\
\dot{\mathbf{q}}&=\frac{1}{2}\boldsymbol{\Omega}\left (\boldsymbol{\omega}_b \right )\mathbf{q}, &\dot{\boldsymbol{\omega}}_b&=\mathbf{J}^{-1}\left ( \boldsymbol{\tau}_\mathrm{r}- \boldsymbol{\omega}_b\times \mathbf{J}\boldsymbol{\omega}_b\right ),
\label{eq:rotational dynamics}
\end{align}
\label{eq:quadrotor model}%
\end{subequations}
where $\mathbf{p}_w=\left [ x,y,z \right ]^{T}$ and $\mathbf{v}_w=\left [ v_x,v_y,v_z \right ]^{T}$ are the position and velocity of the quadrotor's center-of-mass (CoM) in the world frame $\mathcal{W}$, $\mathbf{e}_z=\left [ 0,0,1 \right ]^{T}$, $\mathbf{R}\left ( \mathbf{q} \right )$ is the rotation matrix from the body frame $\mathcal{B}$ to $\mathcal{W}$ parameterized by quaternions $\mathbf{q}=\left [ q_{0},q_{x},q_{y},q_{z} \right ]$, $g$ is the gravitational acceleration, $\boldsymbol{\omega}_b=\left [ {\omega}_x,{\omega}_y,{\omega}_z \right ]^{T}$ is the angular velocity in $\mathcal{B}$, and $\boldsymbol{\Omega}\left (\boldsymbol{\omega}_b \right )$ is a skew-symmetric matrix of $\boldsymbol{\omega}_b$. Finally, $f_\mathrm{r}$ and $\boldsymbol{\tau}_\mathrm{r}=\left [ \tau_x,\tau_y,\tau_z \right ]^{T}$ are the collective thrust and the torque produced by the rotor forces $f_{i},\ \forall i\in \left [ 1,4 \right ]$. We use $\mathbf{x}=\left [ \mathbf{p}_w^{T},\mathbf{v}_w^{T},\mathbf{q}^{T},\boldsymbol{w}_b^{T} \right ]^{T}$ and $\mathbf{u}=\left [ f_1,f_2,f_3,f_4 \right ]^{T}$ to represent the quadrotor's states and control inputs, respectively.

\subsection{Gate Model}\label{subsec:gate kinematics}

\begin{figure}[h]
	\centering
	{\includegraphics[width=0.325\textwidth]{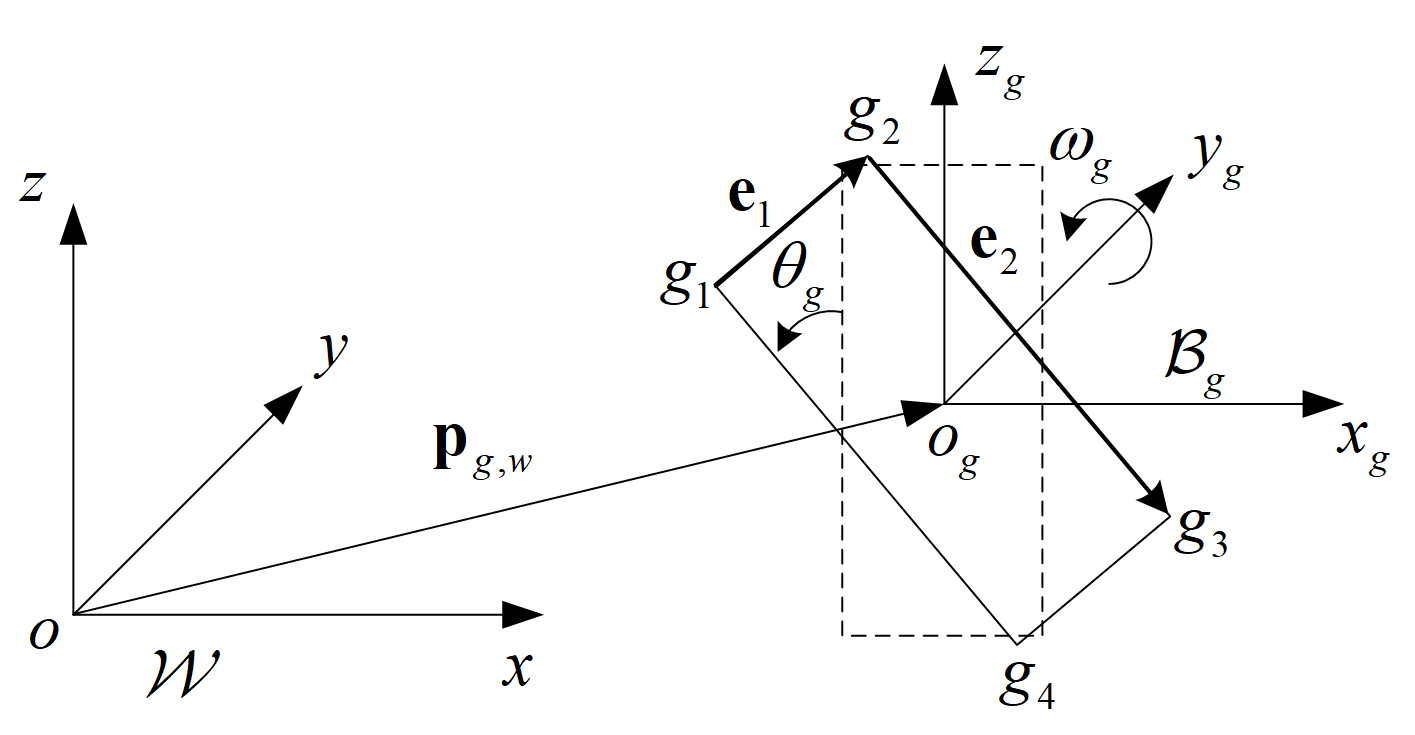}}
	\caption{\footnotesize Illustration of the rectangular gate model and the instantaneously attached body frame $\mathcal{B}_g$. The gate's vertices are denoted by $g_{i=1,2,3,4}$, and $o_g$ is the gate center. For convenience, we define the vectors $\mathbf{e}_{1}$ and $\mathbf{e}_{2}$, which can be expressed in $\mathcal{W}$ in terms of $g_1$, $g_2$, and $g_3$ in $\mathcal{W}$. The axis $\mathbf{y}_{g}$ is not necessarily parallel to $y$ of $\mathcal{W}$.}
\label{fig:gate}	
\end{figure}

We model the dynamic gate as a rectangle which can move freely in space. Fig.\ref{fig:gate} depicts the gate frame $\mathcal{B}_g$ attached to the center and defines the gate on the $x_go_gz_g$ plane. Suppose the gate is perpendicular to the ground and can rotate about its $y_g$ axis freely by a pitch angle $\theta_g$. Note that the gate body frame $\mathcal{B}_g$ is instantaneously attached to the gate at the beginning of movement (the gate in dashed line), which is a non-rotating frame. 
Let $\mathbf{p}_{g,w}=\left [ x_{g},y_{g},z_{g} \right ]^{T}$ denote the position of $o_g$ in $\mathcal{W}$, $\mathbf{v}_{g,w}=\left [ v_{gx},v_{gy},v_{gz} \right ]^{T}$ the velocity of $o_g$ in $\mathcal{W}$, and $\omega_g$ the angular rate. The motion of the dynamic gate is described by:
\begin{equation}
    \dot{\mathbf{p}}_{g,w}=\mathbf{v}_{g,w}, \quad \dot{\theta}_g = \omega_g.
    \label{eq:gate model}
\end{equation}
\subsection{Model Predictive Control}\label{subsec:MPC}
We denote by $\mathbf{x}_{T}$ the quadrotor final states of hovering at a target point behind the gate. MPC can generate an optimal state trajectory $\mathbf{x}_{k}^{\ast },\ \forall k\in \left [ 0,N \right ]$ towards $\mathbf{x}_{T}$ and a sequence of optimal control commands $\mathbf{u}_{k}^{\ast },\ \forall k\in \left [ 0,N-1 \right ]$ over a horizon $N$ that covers the whole flight trajectory. In MPC, we view the gate as an intermediate waypoint and define the cost function as a sum over four quadratic components: a target-tracking cost $J_{x_k}=\left \| \mathbf{x}_{k}-\mathbf{x}_{T} \right \|_{\mathbf{Q}_{x}}^{2}$, two control regularization costs $J_{u_{k}}=\left \| \mathbf{u}_{k} \right \|_{\mathbf{Q}_{u}}^{2}$ and $J_{\Delta u_{k}}=\left \| \mathbf{u}_{k}-\mathbf{u}_{k-1} \right \|_{\mathbf{Q}_{\Delta u}}^{2}$, and a gate-traversing cost $J_{\mathrm{tra},k}=\left \| \boldsymbol{\delta}_{\mathrm{tra},k} \right \|_{\mathbf{Q}_{\mathrm{tra}}}^{2}$ with 
\begin{equation}
    \boldsymbol{\delta}_{\mathrm{tra},k}=\left [\left (\mathbf{p}_{w,k}-\mathbf{p}_{w,\mathrm{tra}}  \right )^{T},\mathrm{Tr}\left ( \mathbf{I}_{3}- \mathbf{R}\left ( \mathbf{q}_{k} \right )^{T}\mathbf{R}\left ( \mathbf{q}_{\mathrm{tra}} \right )\right )  \right ]^{T}
\nonumber
\end{equation}
where $\mathbf{r}_{\mathrm{tra}}=\left [ \mathbf{p}_{w,\mathrm{tra}} ,\mathbf{q}_{\mathrm{tra}}\right ]$ is the reference traversal pose, $\mathbf{I}_{3}\in \mathbb{R}^{3\times 3}$ is an identity matrix, and $\mathrm{Tr}\left ( \cdot  \right )$ takes the trace of a given matrix. Therefore, we solve the following nonlinear optimization problem.
\begin{subequations}
\begin{align}
    \min_{\mathbf{x}_{0:N},\mathbf{u}_{0:N-1}} &\sum_{k=0}^{N-1}\left ( J_{x_k}+J_{u_k}+J_{\Delta u_k} + J_{\mathrm{tra},k} \right )+J_{x_N}\label{eq:mpc cost}\\
    \text{s.t.}\ \mathbf{x}_{k+1}  & =\mathbf{x}_{k}+d_t\ast f_{\mathrm{dyn}}\left ( \mathbf{x}_{k},\mathbf{u}_{k} \right ), \label{eq:mpc dynamics}\\
    \mathbf{x}_{0} &=\mathbf{x}_{\mathrm{init}},\ \mathbf{u}_{-1}=\mathbf{u}_{\mathrm{init}},\  \mathbf{x}_{k}\in \mathbb{X},\ \mathbf{u}_{k}\in \mathbb{U},\label{eq:mpc constraints}
\end{align}
\label{eq:mpc}%
\end{subequations}
where $d_t$ is the discrete time step, $f_{\mathrm{dyn}}$ is the quadrotor's dynamics model defined in (\ref{eq:quadrotor model}), $\mathbf{x}_{\mathrm{init}}$ and $\mathbf{u}_{\mathrm{init}}$ denote the initial states and control commands, $\mathbb{X}$ and $\mathbb{U}$ represent the constraint sets for the quadrotor's states and control inputs, respectively. 

These cost matrices $\mathbf{Q}_x\in \mathbb{R}^{13\times13}$, $\mathbf{Q}_u\in \mathbb{R}^{4\times4}$, $\mathbf{Q}_{{\Delta u}}\in \mathbb{R}^{4\times4}$, and $\mathbf{Q}_{\mathrm{tra}}\in \mathbb{R}^{4\times4}$ are positive definite. In particular, $\mathbf{Q}_x$, $\mathbf{Q}_u$, and $\mathbf{Q}_{{\Delta u}}$ are time-invariant while the traversal cost matrix $\mathbf{Q}_{\mathrm{tra}}$ is time-varying, defined as
\begin{equation}
    \mathbf{Q}_{\mathrm{tra}}\left ( t_{\mathrm{tra}},k \right )=\mathbf{Q}_{\max}\ast \mathrm{exp}\left ( -\gamma \ast \left ( k\ast d_{t}- t_{\mathrm{tra}}\right )^{2} \right ),
    \label{eq:traversal cost matrix}
\end{equation}
where $\mathbf{Q}_{\max}\in \mathbb{R}^{4\times 4}$ is the maximum traversal cost matrix that should be larger than all the time-invariant cost matrices, $\gamma \in \mathbb{R}_{+ }$ is the temporal spread of the traversal cost, and $t_{\mathrm{tra}}$ is the time at which the quadrotor traverses through the gate. As time approaches to $t_{\mathrm{tra}}$, we have $\mathbf{Q}_{\mathrm{tra}}\left ( t_{\mathrm{tra}},k \right )\approx \mathbf{Q}_{\max}$ and the relatively large $\mathbf{Q}_{\max}$ encourages the quadrotor to track the gate. Instead, for time away from $t_{\mathrm{tra}}$, the matrix $\mathbf{Q}_{\mathrm{tra}}$ decays exponentially, making the quadrotor arrive at the target point.

\subsection{Deep Neural SE(3) Decision Variables for MPC}
For Problem (\ref{eq:mpc}), $t_{\mathrm{tra}}$ and $\mathbf{r}_{\mathrm{tra}}$ are of great importance as they determine when, where, and how the quadrotor flies through the gate. Compared to the previous work~\cite{song2022policy}, we learn the more challenging SE(3) traversal references (decision variables) $\mathbf{r}_{\mathrm{tra}}$ instead of merely the traversal position $\mathbf{p}_{w,\mathrm{tra}}$, which significantly improves the feasibility and optimality of SE(3) motion planning. We highlight that the Gaussian sampling based policy search in~\cite{song2022policy} is intractable to learn the SE(3) DNN policy in our case. 

For gradient-based training of the attitude reference, we use the Rodrigues parameters (i.e., the $so(3)$ vector representation) $\boldsymbol\rho_{\mathrm{tra}} \in \mathbb{R}^{3}$ instead of the quaternion $\mathbf{p}_{\mathrm{tra}}$, due to the constraint-free optimization of the former~\cite{diaz20193d}. For ease of reference, we define a vector $\mathbf{z}=\left [ \mathbf{p}_{w,\mathrm{tra}},\boldsymbol\rho_{\mathrm{tra}},t_{\mathrm{tra}} \right ]\in \mathbb{R}^{7}$ as the high-level decision variables, parameterize Problem (\ref{eq:mpc}) as $\mathrm{MPC}\left ( \mathbf{z} \right )$ and denote the corresponding optimal state trajectory by $\boldsymbol{\xi }^{\ast }\left ( \mathbf{z} \right )$ (i.e., $\boldsymbol{\xi }^{\ast }\left ( \mathbf{z} \right )=\left \{\mathbf{x}_{k}^{\ast }\left ( \mathbf{z} \right )  \right \}_{k=0}^{N}$).

In this paper, we are interested in learning deep neural decision variables (i.e., modeling $\mathbf{z}$ using a DNN) that can adapt the MPC performance online in highly dynamic flight scenarios. The training is highly-nontrivial, and we will develop a novel reinforce-imitate learning framework in the following section. It trains the DNN efficiently using a static gate (Section~\ref{sec:learning framework}). Combining with a binary search algorithm, we can update the traversal time predicted by the DNN in real-time to improve the adaptation to dynamic environments (Section~\ref{sec:binary search}). 

\section{Reinforce-imitate Learning Framework}\label{sec:learning framework}
\begin{figure}[h]
	\centering
	{\includegraphics[width=0.37\textwidth]{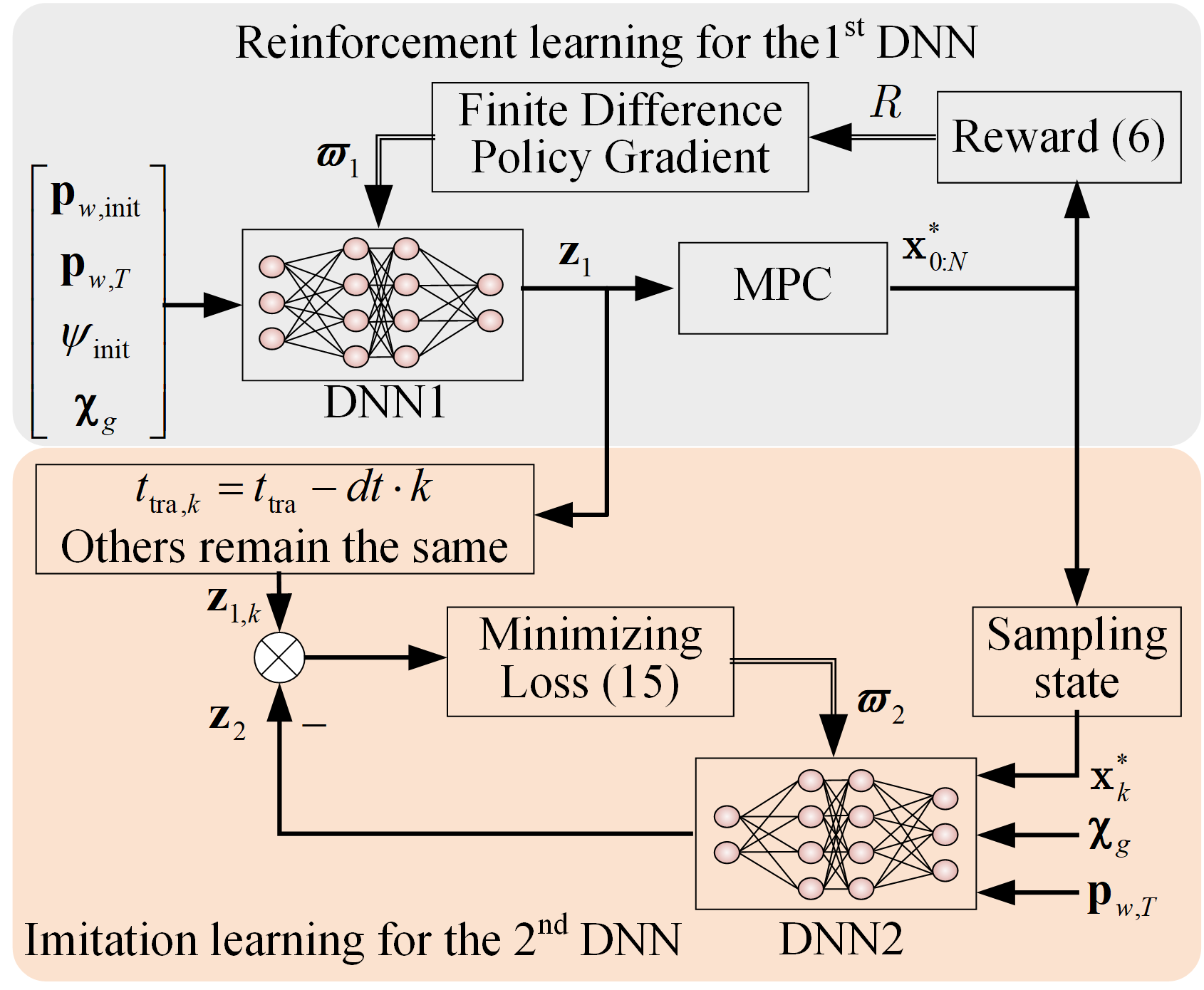}}
	\caption{\footnotesize Diagram of the proposed reinforce-imitate learning framework}
\label{fig:learning pipeline}	
\end{figure}

Fig.\ref{fig:learning pipeline} illustrates the proposed reinforce-imitate learning framework that divides the training process into two tractable subproblems. We train a first DNN via RL to generate the decision variables for a single-run open loop MPC from different stationary states. Then, we train a second DNN via supervised learning to imitate the first while making it adaptive to non-static quadrotor states on the optimal MPC trajectory generated using the first DNN.

\subsection{Reinforcement Learning for Training the 1st DNN}\label{subsec:RL}
We use subscript $1$ to denote the decision variables modeled by the first DNN, which has the following form
\begin{equation}
    \mathbf{z}_1=f_{\boldsymbol{\varpi}_{1} }\left ( \mathbf{p}_{w,\mathrm{init}},\mathbf{p}_{w,T},\psi_{\mathrm{init}},\boldsymbol{\chi }_{g} \right ).
    \label{eq:1st DNN}
\end{equation}
Here $\mathbf{p}_{w,\mathrm{init}}$ and $\psi_{\mathrm{init}}$ denote the initial quadrotor position in $\mathcal{W}$ and yaw angle, $\mathbf{p}_{w,T}$ is the target position in
$\mathcal{W}$, and $\boldsymbol{\varpi}_{1}$ denote the parameters of the first DNN. In training, we place the gate at the origin of $\mathcal{W}$ and use its width $\left \| \mathbf{e}_{1} \right \|_{2}$ and pitch angle $\theta_g$ to denote the gate states $\boldsymbol{\chi}_g$, i.e., $\boldsymbol{\chi}_{g}=\left [ \left \| \mathbf{e}_{1}  \right \|_{2},\theta _{g} \right ]$.

In RL, the reward function is designed as a weighted sum of three terms to evaluate the quality of $\boldsymbol{\xi }^{\ast }\left ( \mathbf{z}_1 \right )$:
\begin{equation}
    R\left ( \boldsymbol{\xi}^{\ast }\left ( \mathbf{z}_1 \right ) \right )=R_{\mathrm{max}}-\alpha  L_{T}\left ( \boldsymbol{\xi}^{\ast }\left ( \mathbf{z}_1 \right ) \right )- \beta L_{\mathrm{coll}}\left ( \boldsymbol{\xi}^{\ast }\left ( \mathbf{z}_1 \right ) \right )
    \label{eq:reward}
\end{equation}
where $\alpha \in \mathbb{R}_{+ }$ and $\beta  \in \mathbb{R}_{+ }$ are the weighting coefficients. We define these three terms and explain their meaning in the following bullets.

\textit{1) Target penalty term $L_{T} \in \mathbb{R}_{+}$:} We introduce this term to penalize the quadrotor's deviation away from the target point. It is defined using the last $n$ successive points on the optimal trajectory, and thus takes the following form: 
\begin{equation}
    L_{T}=\sum_{k=N-n}^{N}\left \| \mathbf{p}_{w,k}^{\ast }\left ( \mathbf{z}_1 \right )-\mathbf{p}_{w,T} \right \|_{2}^{2}.
    \label{eq:deviation penalty}
\end{equation}

\textit{2) Collision penalty term $L_{\mathrm{coll}} \in \mathbb{R}_{+}$:} To penalize the collision, we build this term upon a collision loss function that reflects the traversal performance and can apply to different gate dimensions. However, designing such a function is non-trivial as it requires considering the quadrotor's shape and attitude.
\begin{figure}[h]
	\centering
	{\includegraphics[width=0.35\textwidth]{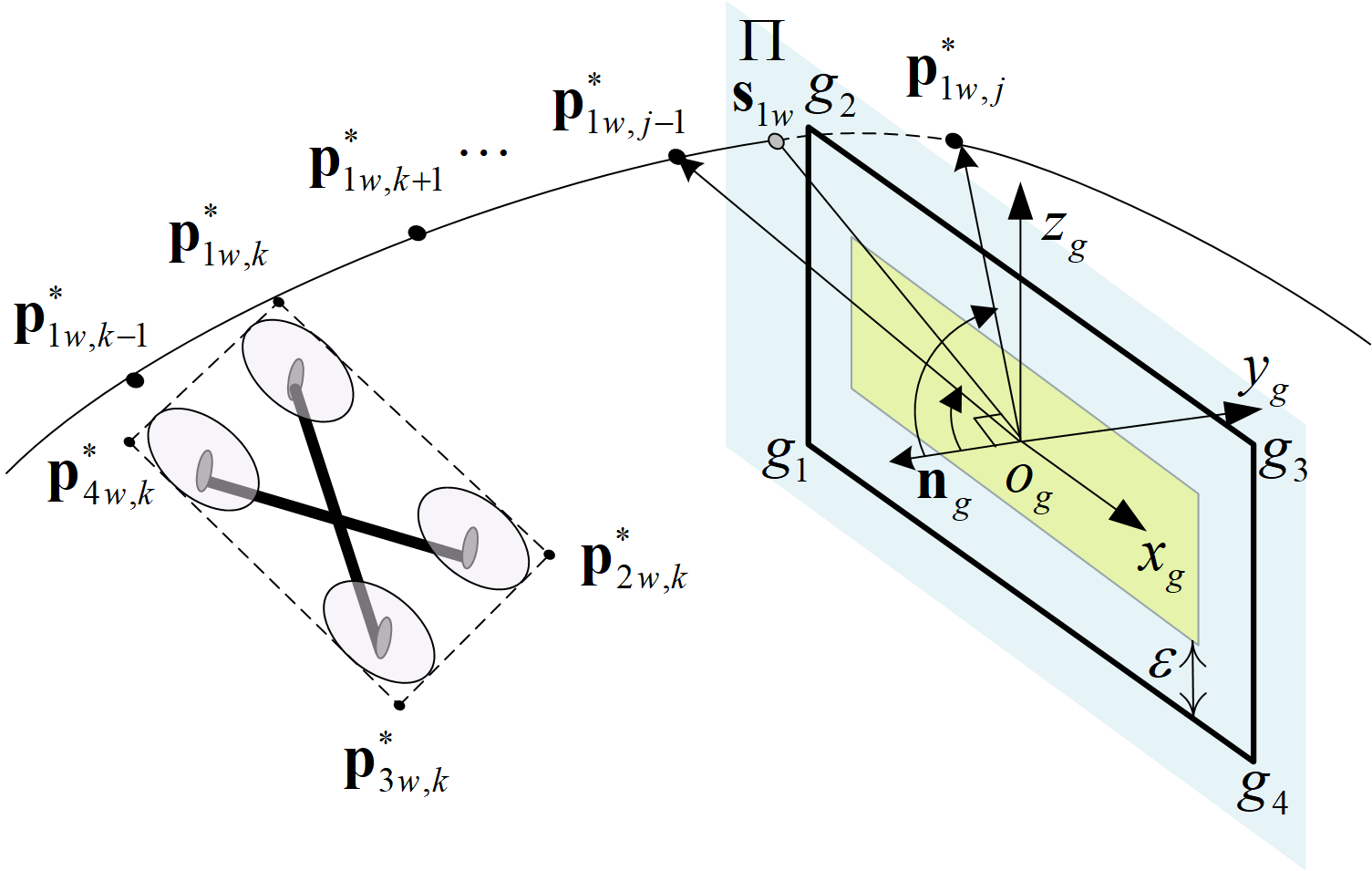}}
	\caption{\footnotesize Given an optimal trajectory of one vertex $\mathbf{p}_{iw,0:N}^{\ast },i\in \left [ 1,4 \right ]$, we compute the inner product of $\mathbf{p}_{iw,k}^{\ast }$ and the normal vector $\mathbf{n}_g$. Suppose $\mathbf{p}_{iw,j}^{\ast }$ is the first waypoint behind the gate that makes the product negative. Then, the line connecting $\mathbf{p}_{iw,j-1}^{\ast }$ and $\mathbf{p}_{iw,j}^{\ast }$ intersects with $\Pi$ at $\mathbf{s}_{iw}$.}
\label{fig:collision detection}	
\end{figure}
Here, we view the quadrotor as a square (denoted by the dashed contour in Fig.\ref{fig:collision detection}) and design the loss functions for the four vertexes. Specifically, given an optimal quadrotor pose $\mathbf{x}_{\mathrm{pose},k}^{\ast}=\left [ \mathbf{p}_{w,k}^{\ast},\mathbf{q}_{k}^{\ast} \right ]$, the position of a vertex in $\mathcal{W}$ at $\mathbf{x}_{\mathrm{pose},k}^{\ast}$ can be obtained as
\begin{equation}
    \mathbf{p}_{iw,k}=\mathbf{p}_{w,k}^{\ast}+\mathbf{R}\left ( \mathbf{q}_{k}^{\ast} \right )\mathbf{p}_{ib,k}, \ k\in \left [ 0,N \right ]
    \label{eq:vertex position}
\end{equation}
where $\mathbf{p}_{ib,k}$ denotes the position of $i$th vertex in $\mathcal{B}$. Then, we compute the intersections $\mathbf{s}_{iw},\ \forall i\in \left [ 1,4 \right ]$ of the vertexes' optimal trajectories and the gate plane $\Pi$, in order to define the collision loss function as:
\begin{equation}
    \mathrm{coll}\left ( \mathbf{s}_{iw} \right )=\left\{\begin{matrix}
\mathrm{max}\left ( 0,\varepsilon -d_i \right) & \mathbf{s}_{iw}\ \mathrm{inside\ the\ gate }\\ 
 2\varepsilon \ast d_i+\varepsilon^{2} & \mathrm{otherwise}
\end{matrix}\right.,
\label{eq:collision loss}
\end{equation}
where $\varepsilon   \in \mathbb{R}_{+ }$ is a safety margin that can be slightly larger than the quadrotor's height, and $d_i$ is the shortest distance from $\mathbf{s}_{iw}$ to the gate sides. This loss function encourages the quadrotor to enter the safe region of the gate (denoted by the light yellow rectangle in Fig.\ref{fig:collision detection}) with a more feasible and optimal pose. This property benefits the motion planning and control after flying through the gate, and thus substantially improves the quadrotor's agility in different environments. Using (\ref{eq:collision loss}), we can design the collision penalty term as
\begin{equation}
    L_{\mathrm{coll}}=\sum_{i=1}^{4}\mathrm{coll}\left ( \mathbf{s}_{iw}\left ( \mathbf{z}_1 \right ) \right ).
    \label{eq:collsion penalty}
\end{equation}

\textit{3) Goal reward term $R_{\mathrm{max}} \in \mathbb{R}_{+}$:} If the quadrotor flies through the gate's safe region and arrives at the target point accurately, this positive goal reward will be achieved.

Hence, the RL problem is to find optimal network parameters $\boldsymbol{\varpi}_{1}^{\ast}$ that maximize the reward $R\left ( \boldsymbol{\xi}^{\ast }\left ( \mathbf{z}_1 \right ) \right )$. It can be interpreted as the following optimization problem:
\begin{subequations}
\begin{align}
\max_{\boldsymbol{\varpi}_{1}}\  &R\left ( \boldsymbol{\xi}^{\ast }\left ( \mathbf{z}_1\left ( \boldsymbol{\varpi}_{1} \right ) \right ) \right )\\
\mathrm{s.t.}\  & \boldsymbol{\xi}^{\ast }\left ( \mathbf{z}_1\left ( \boldsymbol{\varpi}_{1} \right ) \right )\  \mathrm{generated\ by}\ \mathrm{MPC}\left ( \mathbf{z}_1\left ( \boldsymbol{\varpi}_{1} \right ) \right ).
\end{align}
\end{subequations}
We use gradient ascent to train $\boldsymbol{\varpi}_{1}$. The gradient of $R$ with respect to $\boldsymbol{\varpi}_{1}$ can be computed using the chain rule $\frac{dR}{d \boldsymbol{\varpi}_{1}}=\frac{\partial R}{\partial \mathbf{z}_1}\frac{\partial \mathbf{z}_1}{\partial \boldsymbol{\varpi}_{1}}$. Here, calculating $\frac{\partial \mathbf{z}_1}{\partial \boldsymbol{\varpi}_{1}}$ is standard for the neural network via many existing machine learning tools. However, computing $\frac{\partial R}{\partial \mathbf{z}}$ is challenging as it 
requires differentiating through the nonlinear MPC optimization problem (\ref{eq:mpc}), which is computationally expensive for a long horizon. Therefore, we adopt a more efficient method: the finite difference policy gradient. It estimates the gradient $\frac{\partial R}{\partial \mathbf{z}_1}$ by applying a small perturbation $\delta \mathbf{z}_1$ to the decision variable vector $\mathbf{z}_1$, i.e., $\frac{\partial R}{\partial \mathbf{z}}=R\left ( \boldsymbol{\xi}^{\ast }\left ( \mathbf{z}_1+\delta \mathbf{z}_1 \right ) \right )-R\left ( \boldsymbol{\xi}^{\ast }\left ( \mathbf{z}_1 \right ) \right )$. This method is typically powerful in the episode-based learning strategy, provided the reward is not too noisy~\cite{deisenroth2013survey}.

\subsection{Imitation Learning for Training the 2nd DNN}\label{subsec:IL}
We design a second DNN to make the decision variables adaptive to non-static quadrotor states. To this end, the inputs of the second DNN incorporate the current optimal states $\mathbf{x}_{k}^{\ast}$ that move on $\boldsymbol{\xi }^{\ast }\left ( f_{\boldsymbol{\varpi }_1^{\ast}} \right )$, and the resulting decision variables are defined as
\begin{equation}
    \mathbf{z}_2=f_{\boldsymbol{\varpi }_2}\left ( \mathbf{x}_{k}^{\ast},\mathbf{p}_{w,T},\boldsymbol{\chi }_{g} \right ),
    \label{eq:2nd DNN}
\end{equation}
where $\boldsymbol{\varpi}_{2}$ denote the parameters of the second DNN. We train the second DNN to imitate the first via supervised learning. Note that $t_{\mathrm{tra}}$ generated by the first DNN is relative to the initial states. Given the receding horizon manner of MPC, we shift $t_{\mathrm{tra}}$ by the time elapsed till $\mathbf{x}_{k}^{\ast}$, i.e.,
\begin{equation}
    t_{\mathrm{tra},k}=t_{\mathrm{tra}}-k\ast d_{t}.
    \label{eq:time shift}
\end{equation}
The reference traversal pose $\mathbf{r}_{\mathrm{tra}}$ produced by the first DNN keeps the same. Thus, we can denote by 
\begin{equation}
    \mathbf{z}_{1,k}=\left [ \mathbf{p}_{w,\mathrm{tra}},\boldsymbol\rho_{\mathrm{tra}},t_{\mathrm{tra},k} \right ],\ k\in \left [ 0,N \right ],
    \label{eq:desired decision variables for the 2nd DNN}
\end{equation}
the desired decision variables for the second DNN to imitate. The imitation learning problem is to find optimal network parameters $\boldsymbol{\varpi}_{2}^{\ast}$ that minimize the following loss:
\begin{equation}
    L=\left \| f_{\boldsymbol{\varpi }_{2}}\left ( \mathbf{x}_{k}^{\ast},\mathbf{p}_{w,T},\boldsymbol{\chi }_{g} \right )-\mathbf{z}_{1,k} \right \|_{2}^{2}.
    \label{eq:mse loss}
\end{equation}
We summarize the whole training procedures of the proposed reinforce-imitate learning framework in Algorithm~\ref{alg:learning framework}.

\begin{algorithm}[!h]
\caption{Reinforce-Imitate Learning Framework}
\label{alg:learning framework}
\begin{algorithmic}[1]
\State \textbf{Input} $f_{\boldsymbol{\varpi }_{1}}$
\Repeat{ Reinforcement learning for the 1st DNN} 
\State Randomly sample $\mathbf{p}_{w,\mathrm{init}}$, $\mathbf{p}_{w,T}$, $\psi_{\mathrm{init}}$, and $\boldsymbol{\chi }_{g}$
\State Compute $\mathbf{z}_1=f_{\boldsymbol{\varpi}_{1} }\left ( \mathbf{p}_{w,\mathrm{init}},\mathbf{p}_{w,T},\psi_{\mathrm{init}},\boldsymbol{\chi }_{g} \right )$
\State Solve $\mathrm{MPC}\left ( \mathbf{z}_1 \right )$ to obtain $\boldsymbol{\xi}^{\ast }\left ( \mathbf{z}_1 \right )$ 
\State Solve $\mathrm{MPC}\left ( \mathbf{z}_1+\delta \mathbf{z}_1 \right )$ to obtain $\boldsymbol{\xi}^{\ast }\left ( \mathbf{z}_1+\delta \mathbf{z}_1 \right )$
\State Compute $\frac{\partial R}{\partial \mathbf{z}_1}=R\left ( \boldsymbol{\xi}^{\ast }\left ( \mathbf{z}_1+\delta \mathbf{z}_1 \right ) \right )-R\left ( \boldsymbol{\xi}^{\ast }\left ( \mathbf{z}_1 \right ) \right )$
\State Update $\boldsymbol{\varpi }_{1}$ using gradient-based optimization
\Until{$f_{\boldsymbol{\varpi }_{1}}$ is well trained}
\State \textbf{Output} $f_{\boldsymbol{\varpi }_{1}^{\ast}}$
\State \textbf{Input} $f_{\boldsymbol{\varpi }_{2}}$
\Repeat{ Imitation learning for the 2nd DNN} 
\State Randomly sample $\mathbf{p}_{w,{\mathrm{init}}}$, $\mathbf{p}_{w,T}$, $\psi_{\mathrm{init}}$, and $\boldsymbol{\chi }_{g}$
\State Compute $\mathbf{z}_1=f_{\boldsymbol{\varpi}_{1}^{\ast} }\left ( \mathbf{p}_{w,{\mathrm{init}}},\mathbf{p}_{w,T},\psi_{\mathrm{init}},\boldsymbol{\chi }_{g} \right )$
\State Solve $\mathrm{MPC}\left ( \mathbf{z}_1 \right )$ to obtain $\boldsymbol{\xi}^{\ast }\left ( \mathbf{z}_1 \right )$
\For{$k \gets 0$ to $N$}
\State Sample $\mathbf{x}_{k}^{\ast }$ from $\boldsymbol{\xi}^{\ast }\left ( \mathbf{z}_1 \right )$
\State Compute $t_{\mathrm{tra},k}=t_{\mathrm{tra}}-k\ast d_{t}$ to obtain $\mathbf{z}_{1,k}$
\State Compute $L=\left \| f_{\boldsymbol{\varpi }_{2}}\left ( \mathbf{x}_{k}^{\ast},\mathbf{p}_{w,T},\boldsymbol{\chi }_{g} \right )-\mathbf{z}_{1,k} \right \|_{2}^{2}$
\State Update $\boldsymbol{\varpi }_{2}$ using gradient-based optimization
\EndFor
\Until{$f_{\boldsymbol{\varpi }_{2}}$ is well trained}
\State \textbf{Output} $f_{\boldsymbol{\varpi }_{2}^{\ast}}$
\end{algorithmic}
\end{algorithm}

\section{Binary Search for Traversal Time}\label{sec:binary search}

\begin{figure}[h]
	\centering
	{\includegraphics[width=0.35\textwidth]{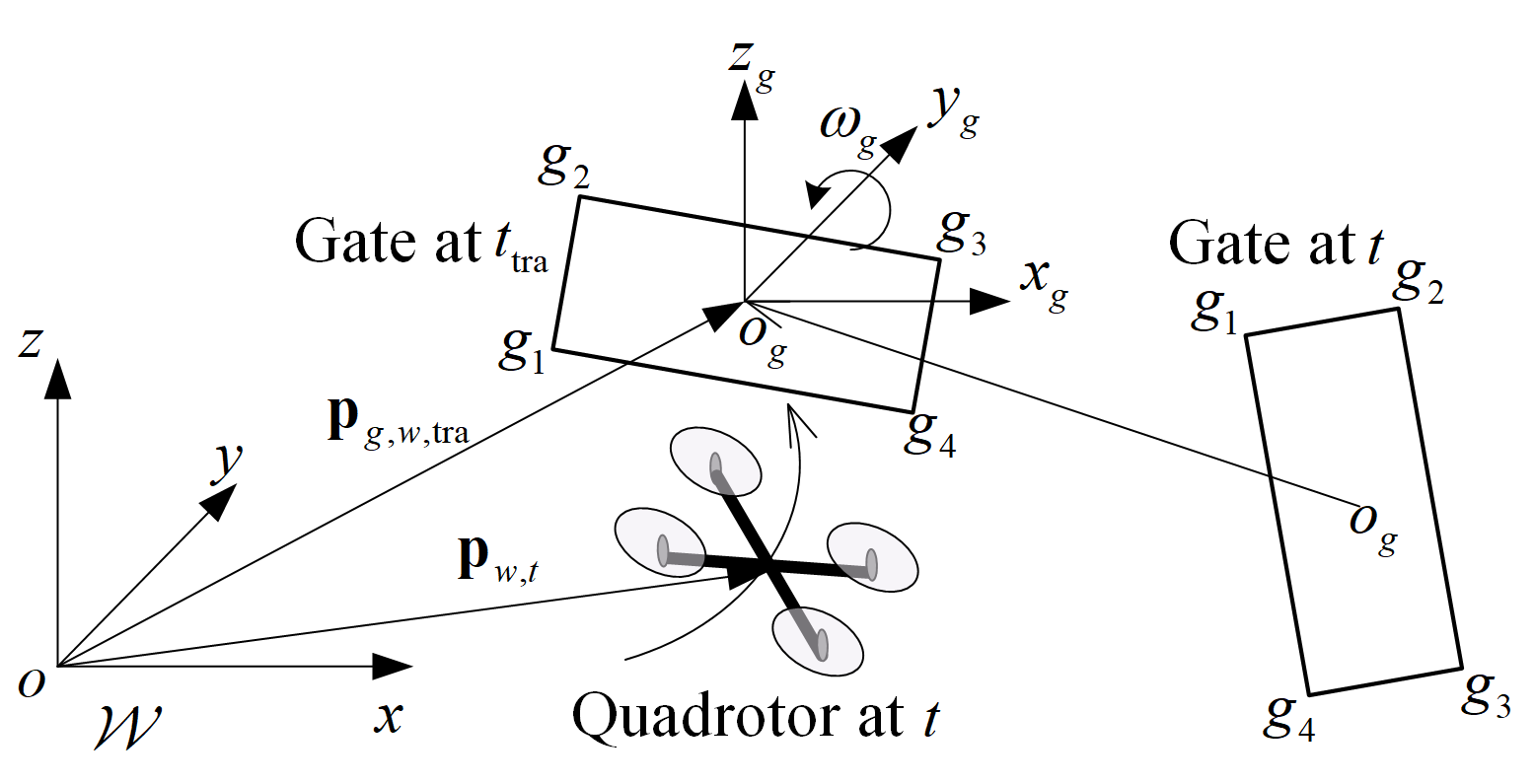}}
	\caption{\footnotesize A dynamic gate viewed as "static" at its traversal position.}
\label{fig:binary_search}	
\end{figure}

We further design a binary search method that efficiently applies the second DNN to a dynamic gate. The idea is to view the gate as "static" at its traversal position (denoted by 'gate at $t_{\mathrm{tra}}$' in Fig.\ref{fig:binary_search}) 
and transform the network inputs\footnote{When using the binary search method, we solve the MPC in the receding horizon manner and replace $\mathbf{x}_{k}^{\ast}$ in (\ref{eq:2nd DNN}) with the current quadrotor states $\mathbf{x}_t$.} to the gate frame of that position (see Fig.\ref{fig:overview}). Such a transformation of the inputs enables the second DNN to generate the corresponding decision variables that guide the quadrotor towards the "static" gate. 

Using the gate kinematics model (\ref{eq:gate model}), we can predict the gate traversal position by updating $t_{\mathrm{tra}}$ from the second DNN via the binary search method. We start with an initial guess $t_1$, which is proportional to the distance between the current quadrotor's CoM and the gate center, and calculate the future gate position at $t_1$. Then, the network inputs are transformed into the gate frame at $t_1$ and thus the second DNN can generate the corresponding traversal time\footnote{To avoid confusion with the actual traversal time $t_{\mathrm{tra}}$, we denote by $t_2$ the traversal time from the second DNN in the binary search method.} $t_2$. Finally, we update $t_1$ using the mean value of these two times. We repeat these steps until the difference between $t_1$ and $t_2$ is within a small threshold $\epsilon$, meaning that both the times converge to the actual traversal time $t_{\mathrm{tra}}$. The whole procedures of the binary search method are summarized in Algorithm~\ref{alg:binary search}. As a consequence of the well-trained second DNN, this algorithm can be run in real-time to substantially improve the performance of tracking a dynamic gate.
\begin{algorithm}[!h]
\caption{Binary Search Algorithm for Updating $t_{\mathrm{tra}}$}
\label{alg:binary search}
\begin{algorithmic}[1]
\State \textbf{Input} $f_{\boldsymbol{\varpi }_{2}^{\ast}}$, $\mathbf{x}$, $\mathbf{v}_{g,w}$, and $\omega_g$
\State Guess an initial value of $t_1$ and set $t_{2} = 0$
\While{$\left | t_{1}-t_{2} \right |> \epsilon $}
\State Predict $\mathbf{p}_{g,w}$ and $\theta_g$ of the gate using Eq.(\ref{eq:gate model}) and $t_1$
\State Transform $\mathbf{x}_t$ and $\mathbf{p}_{w,T}$ into the gate frame at $t_1$
\State Obtain $t_2$ from the second DNN $f_{\boldsymbol{\varpi }_2^{\ast}}$
\State Update $t_{1}\gets\frac{1}{2}\left ( t_{1}+t_{2} \right )$
\EndWhile
\end{algorithmic}
\end{algorithm}

\section{Simulation Results}\label{sec:experiment}
We validate the effectiveness of our approach via extensive simulations. In particular, we design a challenging scenario: traversing through a fast-moving and rotating gate to reach a random target point behind it. The gate moves from the origin of $\mathcal{W}$ at a time-varying velocity of $\mathbf{v}_{g,w}\sim \mathcal{N}\left ( \boldsymbol{\mu},\boldsymbol{\sigma} \right )\ \mathrm{m}\cdot \mathrm{s}^{-1}$ where $\boldsymbol{\mu}$ is the gate mean velocity and $\boldsymbol{\sigma}$ is the gate velocity standard deviation, while rotating at a constant angular rate of $\frac{\pi}{2}\ \mathrm{rad}\cdot \mathrm{s}^{-1}$ from the initial angle $\theta_{g,\mathrm{init}}$ that is inversely proportional to the gate width. We place the target around $\mathbf{p}_{w,T\mathrm{c}}=\left [ 0,6,0 \right ]^{T} \ \mathrm{m}$ subject to an uniform random deviation $\Delta \mathbf{p}_{w,T}\sim \mathcal{U}_{\left [ -\mathbf{2},{\mathbf{2}} \right ]}\ \mathrm{m}$ where $\mathbf{2}=\left [ 2,2,2 \right ]$, such that $\mathbf{p}_{w,T}=\mathbf{p}_{w,T\mathrm{c}}+\Delta \mathbf{p}_{w,T}$. The quadrotor's position and yaw angle are randomly initialized by $\mathbf{p}_{w,\mathrm{init}}=\mathbf{p}_{w,\mathrm{init}c}+\Delta \mathbf{p}_{w,\mathrm{init}}$ and $\psi_{\mathrm{init}}\sim \mathcal{U}_{\left [ -0.1,0.1 \right ]} \ \mathrm{rad}$, respectively, where $\mathbf{p}_{w,\mathrm{init}c}=\left [ 0,-9,0 \right ]^{T}\ \mathrm{m}$ and $\Delta \mathbf{p}_{w,\mathrm{init}}\sim \mathcal{U}_{\left [ -\mathbf{5},\mathbf{5} \right ]}\ \mathrm{m}$. The rest of the quadrotor initial states are fixed at zeros. For better visualization in 3D simulation, we scale up the gate's length to $2 \ \mathrm{m}$, and its width is sampled from a Gaussian distribution, i.e., $\left \| \mathbf{e}_{1} \right \|_{2}\sim \mathcal{N}\left ( 0.9,0.2 \right )\ \mathrm{m} $, in the beginning of each flight. Accordingly, the quadrotor's dimension is scaled up to $1.5 \ \mathrm{m} \times 1.5 \ \mathrm{m}$, but not affecting its dynamics properties.

In the MPC, we use a prediction time horizon of $T=5.0\ \mathrm{s}$ and a discrete time step of $d_{t}=0.1 \ \mathrm{s}$. In the simulation environment, the quadrotor model (\ref{eq:quadrotor model}) and the gate model (\ref{eq:gate model}) are integrated using the forward Euler method with the discrete time step of $0.01\ \mathrm{s}$. To build those two DNNs, we adopt two multilayer perceptrons (MLPs) with rectified linear unit (ReLU) as the activation function. The architectures of the first and second DNNs are $9\rightarrow 64\rightarrow 64\rightarrow 7$ and $18\rightarrow 128\rightarrow 128\rightarrow 7$, respectively. We use CasADi~\cite{andersson2019casadi} with \texttt{ipopt} to solve the MPC optimization problem (\ref{eq:mpc}) in Python. These two MLPs are built in PyTorch~\cite{paszke2019pytorch} and are trained using \texttt{Adam}~\cite{kingma2014adam}.

\begin{figure}[h]
	\centering
	{\includegraphics[width=0.45\textwidth]{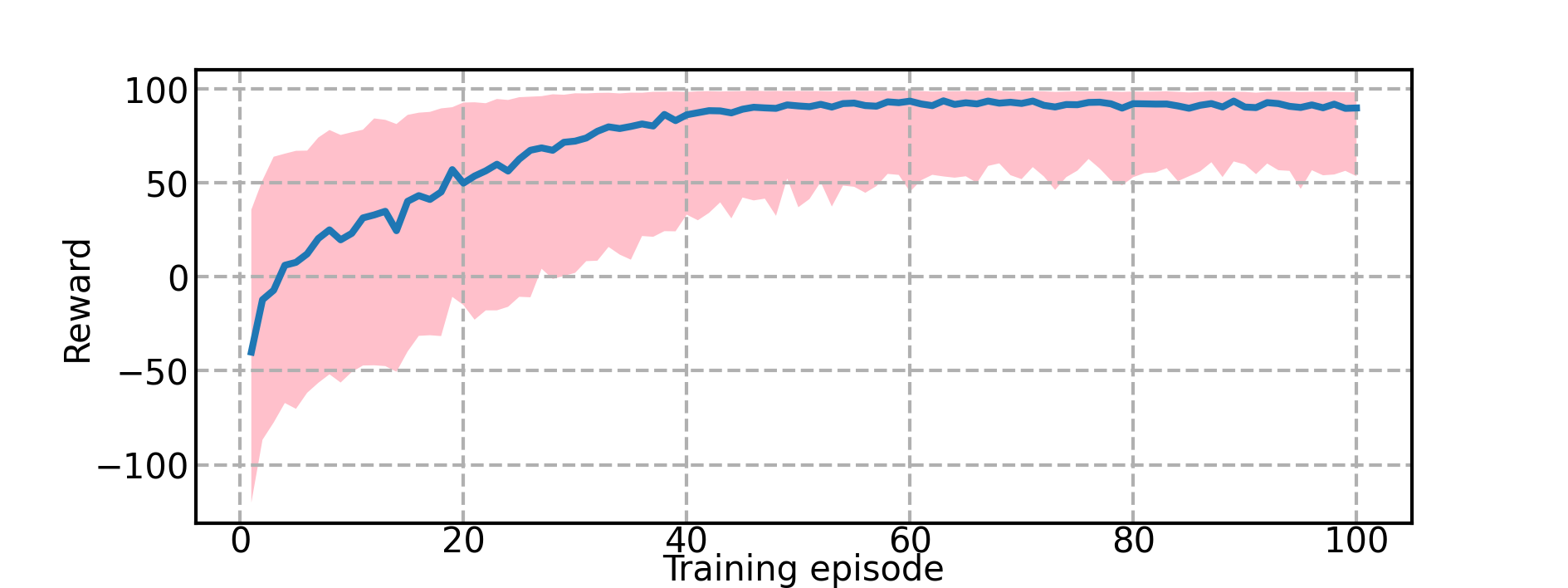}}
	\caption{\footnotesize Reward of training the first DNN. The goal reward $R_{\max}$ is $100$. In each episode, we iteratively solve the MPC $100$ times. We compute the median value (solid curve) and the interquartile range (pink region). Here we set the safe margin $\varepsilon$ to be $0.2\ \mathrm{m}$. A larger $\varepsilon$ can reduce the collision risk but escalates the training difficulty, leading to a smaller steady reward.}
\label{fig:reward}	
\end{figure}

\begin{figure*}[t]
\centering
\begin{subfigure}[b]{0.49\textwidth}
\centering
\includegraphics[width=0.88\textwidth]{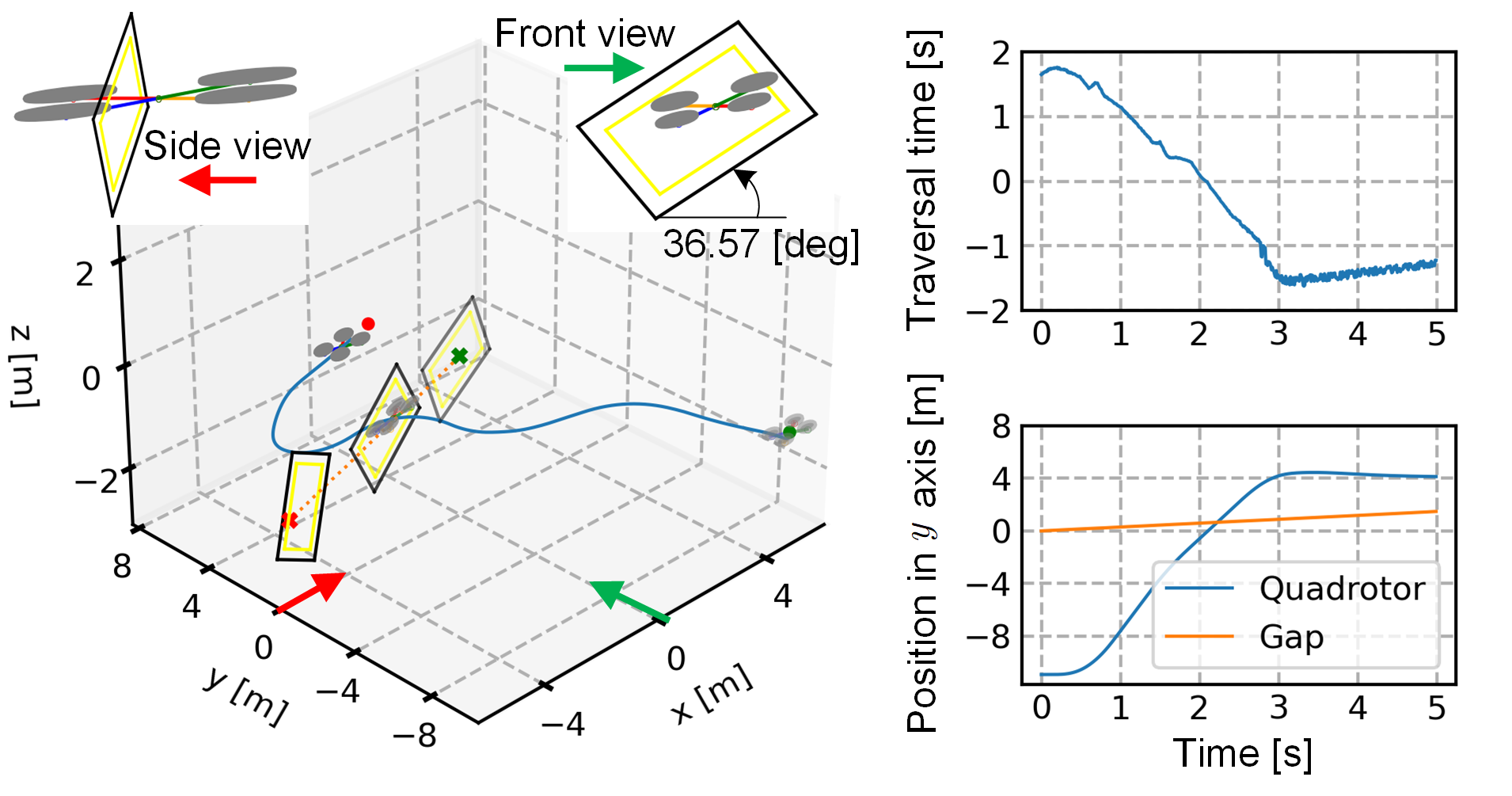}
\caption{\footnotesize Trial 1 with $\boldsymbol{\mu}=\left[ -1,0.3,-0.4 \right]^{T} \mathrm{m}\cdot \mathrm{s}^{-1}$}
\label{fig:left down}
\end{subfigure}
\hfill
\begin{subfigure}[b]{0.49\textwidth}
\centering
\includegraphics[width=0.88\textwidth]{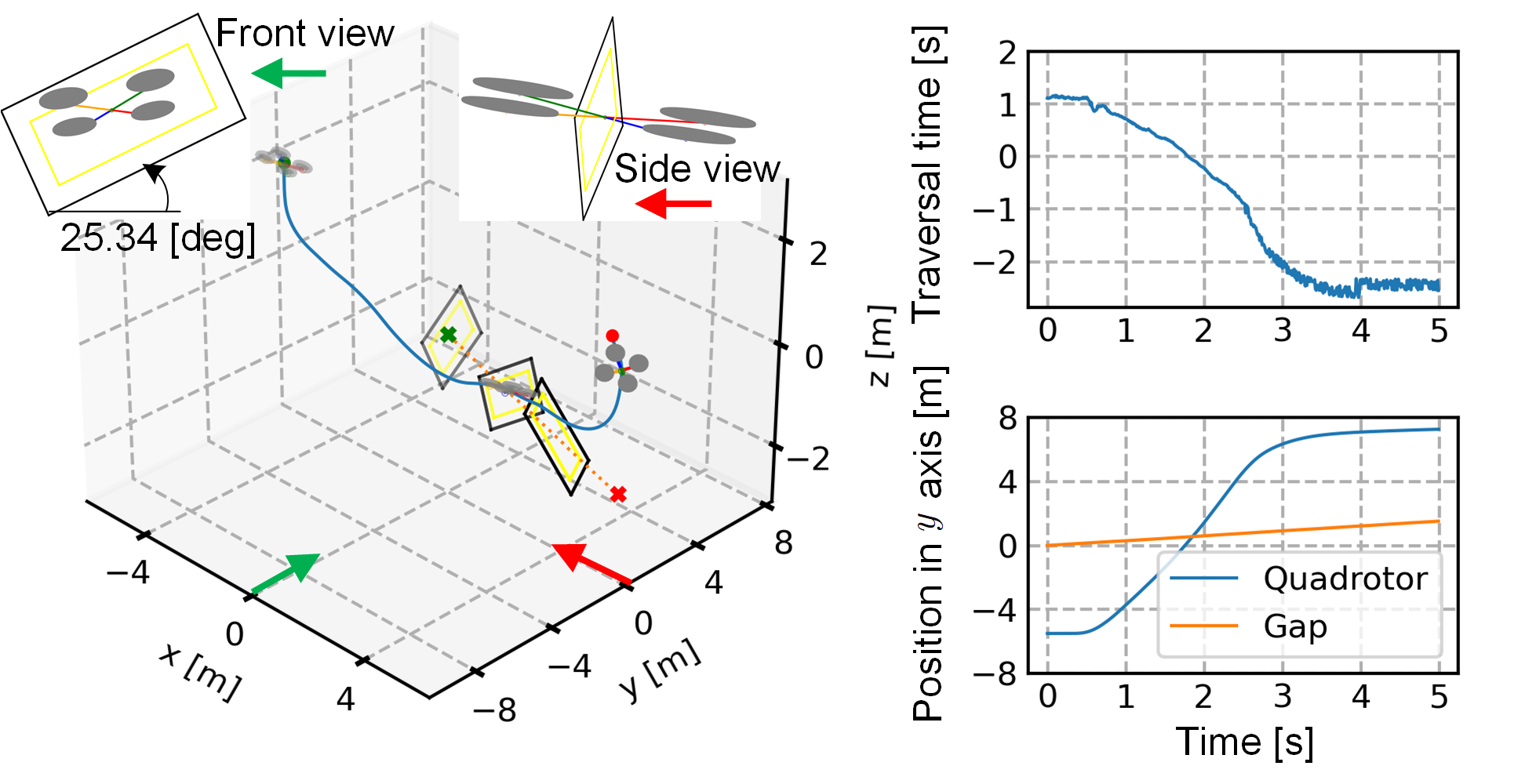}
\caption{\footnotesize Trial 2 with $\boldsymbol{\mu}=\left[ 1,0.3,-0.4 \right]^{T}\mathrm{m}\cdot \mathrm{s}^{-1}$}
\label{fig:right down}
\end{subfigure}
\vskip\baselineskip
\begin{subfigure}[b]{0.49\textwidth}
\centering
\includegraphics[width=0.88\textwidth]{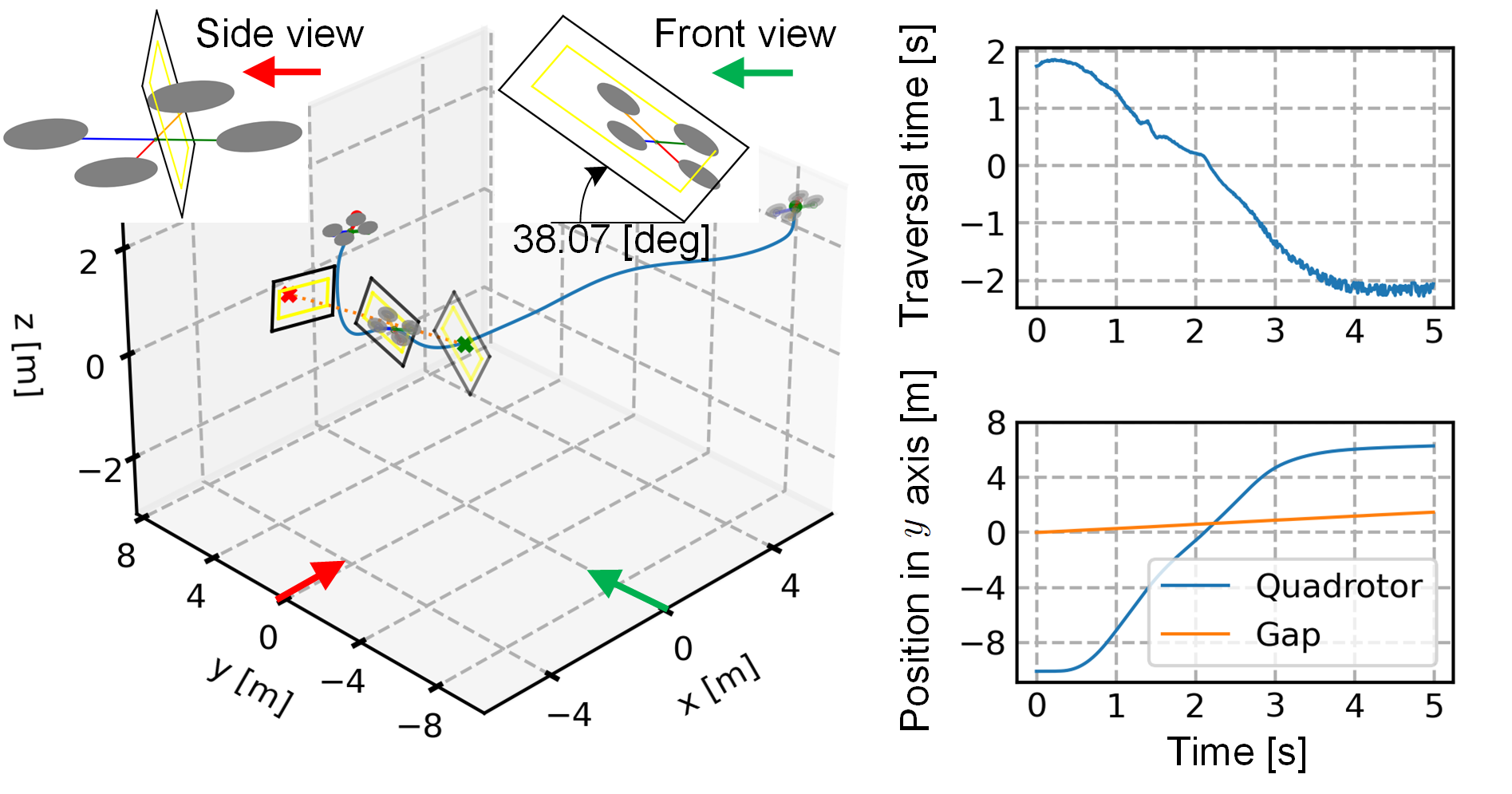}
\caption{\footnotesize Trial 3 with $\boldsymbol{\mu}=\left[ -1,0.3,0.4 \right]^{T}\mathrm{m}\cdot \mathrm{s}^{-1}$}
\label{fig:left up}
\end{subfigure}
\hfill
\begin{subfigure}[b]{0.49\textwidth}
\centering
\includegraphics[width=0.88\textwidth]{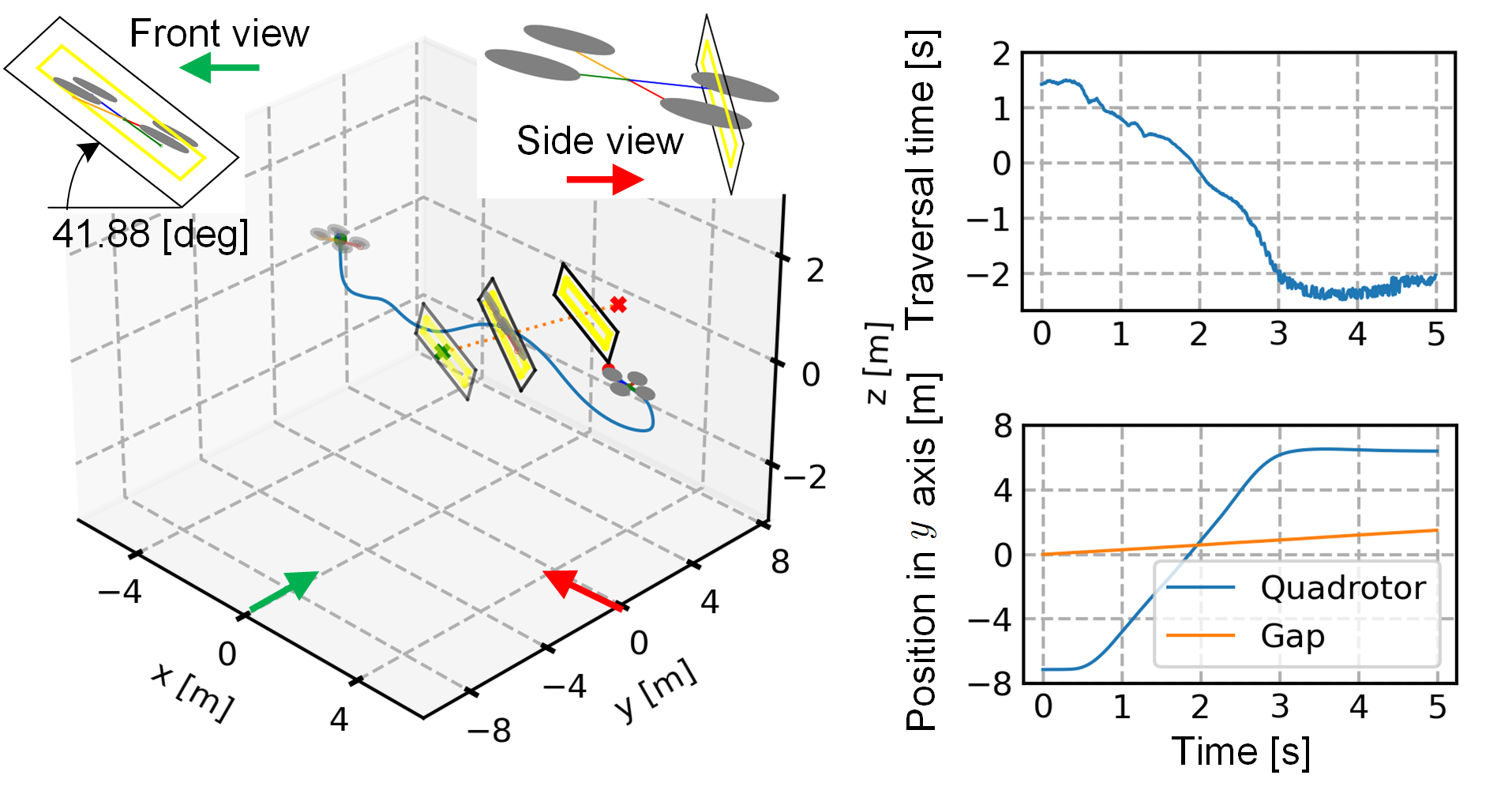}
\caption{\footnotesize Trial 4 with $\boldsymbol{\mu}=\left[ 1,0.3,0.4 \right]^{T}\mathrm{m}\cdot \mathrm{s}^{-1}$}
\label{fig:right up}
\end{subfigure}
\caption{\footnotesize Evaluation of the trained second DNN and the binary search algorithm with $\boldsymbol{\sigma}=\mathrm{diag}\left ( 0.1,0.1,0.1 \right )\ \mathrm{m}\cdot \mathrm{s}^{-1}$. The green and red dots represent the start and target points of the quadrotor, respectively, while the green and red crosses denote the initial and final positions of the gate. The yellow rectangles in the zoom-in sub-figures denote the safe regions. }
\label{fig:visualization}%
\end{figure*}

Fig.\ref{fig:reward} shows the reward $R\left ( \boldsymbol{\xi}^{\ast }\left ( \mathbf{z}_1 \right ) \right )$ during the training of the first DNN via the RL of Algorithm~\ref{alg:learning framework}.
The median reward increases rapidly in the beginning and converges to around $90$ after $50$ episodes, indicating efficient and stable learning. Based on the first DNN, we then train the second DNN using the imitation learning of Algorithm~\ref{alg:learning framework}.

We evaluate the trained second DNN and Algorithm~\ref{alg:binary search} on a dynamic gate whose velocity profile is defined above. In evaluation, the MPC is implemented in the receding horizon manner where we set $\mathbf{x}_{\mathrm{init}}=\mathbf{x}_{t}$ and $\mathbf{u}_{\mathrm{init}}=\mathbf{u}_{t-1}$, and apply $\mathbf{u}_{0}^{\ast}$ to the quadrotor. We visualize four randomly sampled examples with different gate mean velocities in Fig.\ref{fig:visualization}. Our method enables the quadrotor to fly through the dynamic gate and approach the target as closely as possible, adapting to different gate orientations, trajectories, and dimensions. Specifically, the quadrotor can enter the pre-defined safe region of the gate with an optimal pose (see the zoom-in sub-figures). We further observe from the right columns in Fig.\ref{fig:visualization} that the positions of the quadrotor and the gate in the $y$ direction intersect when the traversal time decreases to zero (see Eq.(\ref{eq:time shift}) for the decreasing $t_{\mathrm{tra}}$), indicating accurate updates of $t_{\mathrm{tra}}$ by the developed binary search algorithm.

\begin{figure}[h]
	\centering
	{\includegraphics[width=0.45\textwidth]{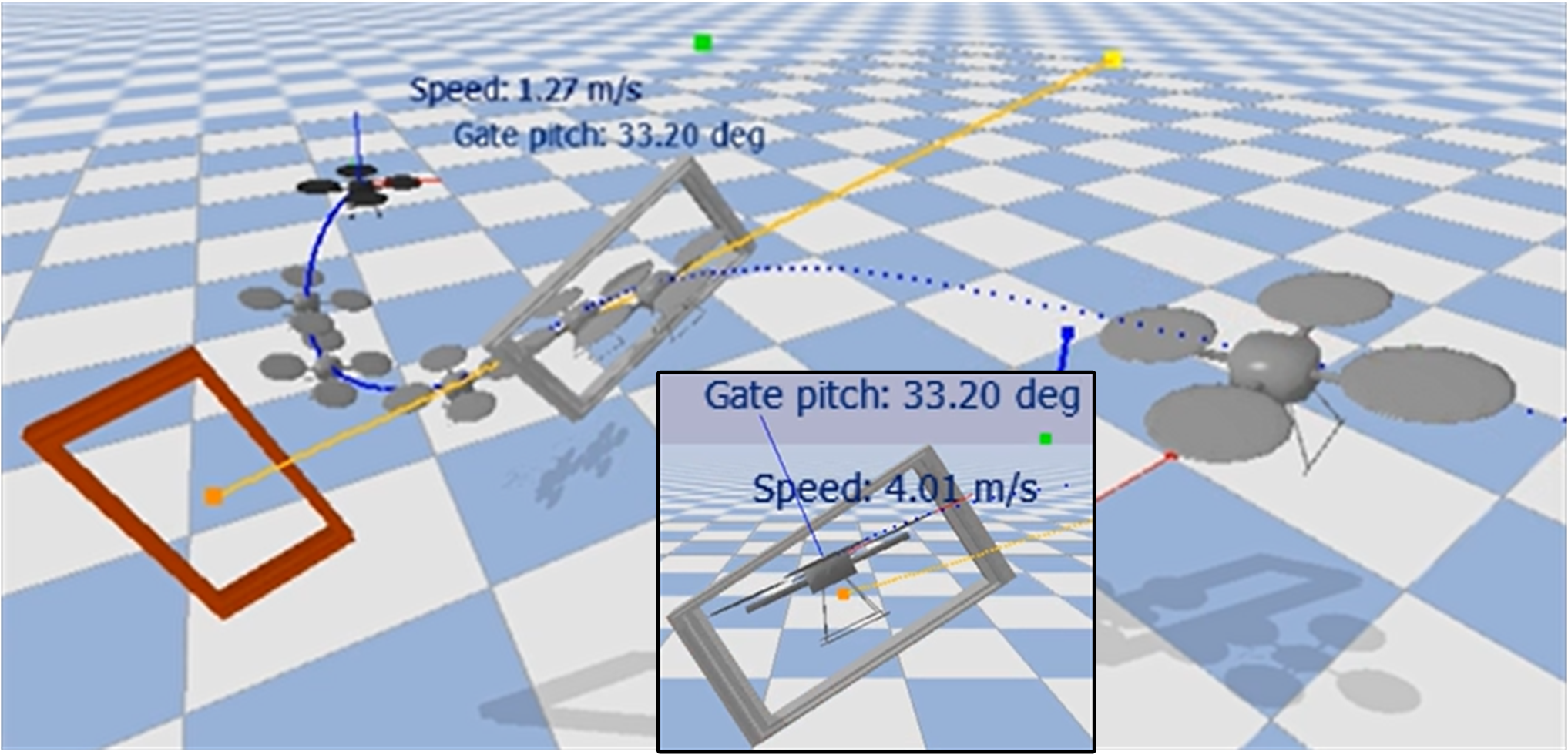}}
	\caption{\footnotesize A trial of flying through a fast-moving and rotating gate in the \texttt{gym-pybullet-drones}~\cite{panerati2021learning}. The zoom-in figure shows the quadrotor traversing through the tilted gate of $33.2 \ \mathrm{deg}$ with an optimal pose.}
\label{fig:pybullet}	
\end{figure}

We further test our method in \texttt{gym-pybullet-drones}, a high-fidelity open-source quadrotor simulator considering more complex aerodynamic effects than our training settings, such as drag and ground effect~\cite{panerati2021learning}. The MPC and the trained second DNN are deployed directly to the simulator without extra tuning. We randomly run four trials using the same gate mean velocities as those in Fig.\ref{fig:visualization}. Fig.\ref{fig:pybullet} visualizes a trial corresponding to $\boldsymbol{\mu} = \left[ -1,0.3,-0.4 \right]^{T}\  \mathrm{m}\cdot \mathrm{s}^{-1}$, and the complete demo can be found in this video: \url{https://youtu.be/TCId8S22gic}. These successes demonstrate our approach's robustness to aerodynamic disturbances and great potential for practical applications.

\section{Conclusions}\label{sec:conclusion} 
Driven by the challenging gate-traversing agile flight, this paper presented a novel deep SE(3) motion planning and control method for quadrotors. It learns an MPC's adaptive SE(3) decision variables parameterized by a portable DNN, encouraging the quadrotor to fly through the gate with maximum safety margins under diverse settings. Our reinforce-imitate learning framework and binary search algorithm allow efficient training with a static gate and then adaptation to highly dynamic environments. Extensive high-fidelity simulations suggest that our method is adaptive to different gate trajectories, velocities, orientations, and target positions. Our future work includes theoretical studies on the robustness and stability of the deep SE(3) MPC control framework. We also plan to develop more efficient training algorithms on SE(3) with analytical gradients.

\addtolength{\textheight}{-12cm}   





\bibliographystyle{IEEEtran}
\bibliography{reference}

@inproceedings{mellinger2011minimum,
  title={Minimum snap trajectory generation and control for quadrotors},
  author={Mellinger, Daniel and Kumar, Vijay},
  booktitle={2011 IEEE international conference on robotics and automation (ICRA)},
  pages={2520--2525},
  year={2011},
  organization={IEEE}
}

@article{wang2022geometrically,
  title={Geometrically constrained trajectory optimization for multicopters},
  author={Wang, Zhepei and Zhou, Xin and Xu, Chao and Gao, Fei},
  journal={IEEE Transactions on Robotics},
  year={2022},
  publisher={IEEE}
}

@inproceedings{panerati2021learning,
  title={Learning to fly—a gym environment with pybullet physics for reinforcement learning of multi-agent quadcopter control},
  author={Panerati, Jacopo and Zheng, Hehui and Zhou, SiQi and Xu, James and Prorok, Amanda and Schoellig, Angela P},
  booktitle={2021 IEEE/RSJ International Conference on Intelligent Robots and Systems (IROS)},
  pages={7512--7519},
  year={2021},
  organization={IEEE}
}

@inproceedings{shim2003decentralized,
  title={Decentralized nonlinear model predictive control of multiple flying robots},
  author={Shim, David Hyunchul and Kim, H Jin and Sastry, Shankar},
  booktitle={42nd IEEE International Conference on Decision and Control (IEEE Cat. No. 03CH37475)},
  volume={4},
  pages={3621--3626},
  year={2003},
  organization={IEEE}
}

@article{han2021fast,
  title={Fast-Racing: An Open-Source Strong Baseline for {SE}(3) Planning in Autonomous Drone Racing},
  author={Han, Zhichao and Wang, Zhepei and Pan, Neng and Lin, Yi and Xu, Chao and Gao, Fei},
  journal={IEEE Robotics and Automation Letters},
  volume={6},
  number={4},
  pages={8631--8638},
  year={2021},
  publisher={IEEE}
}

@inproceedings{lee2010geometric,
  title={Geometric tracking control of a quadrotor UAV on {SE}(3)},
  author={Lee, Taeyoung and Leok, Melvin and McClamroch, N Harris},
  booktitle={49th IEEE conference on decision and control (CDC)},
  pages={5420--5425},
  year={2010},
  organization={IEEE}
}

@article{pereira2021nonlinear,
  title={Nonlinear model predictive control on {SE}(3) for quadrotor aggressive maneuvers},
  author={Pereira, Jean C and Leite, Valter JS and Raffo, Guilherme V},
  journal={Journal of Intelligent \& Robotic Systems},
  volume={101},
  number={3},
  pages={1--15},
  year={2021},
  publisher={Springer}
}

@article{mueller2015computationally,
  title={A computationally efficient motion primitive for quadrocopter trajectory generation},
  author={Mueller, Mark W and Hehn, Markus and D'Andrea, Raffaello},
  journal={IEEE transactions on robotics},
  volume={31},
  number={6},
  pages={1294--1310},
  year={2015},
  publisher={IEEE}
}

@article{mellinger2012trajectory,
  title={Trajectory generation and control for precise aggressive maneuvers with quadrotors},
  author={Mellinger, Daniel and Michael, Nathan and Kumar, Vijay},
  journal={The International Journal of Robotics Research},
  volume={31},
  number={5},
  pages={664--674},
  year={2012},
  publisher={SAGE Publications Sage UK: London, England}
}

@article{loianno2016estimation,
  title={Estimation, control, and planning for aggressive flight with a small quadrotor with a single camera and {IMU}},
  author={Loianno, Giuseppe and Brunner, Chris and McGrath, Gary and Kumar, Vijay},
  journal={IEEE Robotics and Automation Letters},
  volume={2},
  number={2},
  pages={404--411},
  year={2016},
  publisher={IEEE}
}

@inproceedings{falanga2017aggressive,
  title={Aggressive quadrotor flight through narrow gaps with onboard sensing and computing using active vision},
  author={Falanga, Davide and Mueggler, Elias and Faessler, Matthias and Scaramuzza, Davide},
  booktitle={2017 IEEE international conference on robotics and automation (ICRA)},
  pages={5774--5781},
  year={2017},
  organization={IEEE}
}

@article{liu2018search,
  title={Search-based motion planning for aggressive flight in {SE} (3)},
  author={Liu, Sikang and Mohta, Kartik and Atanasov, Nikolay and Kumar, Vijay},
  journal={IEEE Robotics and Automation Letters},
  volume={3},
  number={3},
  pages={2439--2446},
  year={2018},
  publisher={IEEE}
}

@inproceedings{lin2019flying,
  title={Flying through a narrow gap using neural network: an end-to-end planning and control approach},
  author={Lin, Jiarong and Wang, Luqi and Gao, Fei and Shen, Shaojie and Zhang, Fu},
  booktitle={2019 IEEE/RSJ International Conference on Intelligent Robots and Systems (IROS)},
  pages={3526--3533},
  year={2019},
  organization={IEEE}
}

@article{xiao2021flying,
  title={Flying Through a Narrow Gap Using End-to-End Deep Reinforcement Learning Augmented With Curriculum Learning and Sim2Real},
  author={Xiao, Chenxi and Lu, Peng and He, Qizhi},
  journal={IEEE Transactions on Neural Networks and Learning Systems},
  year={2021},
  publisher={IEEE}
}

@article{song2022policy,
  title={Policy Search for Model Predictive Control With Application to Agile Drone Flight},
  author={Song, Yunlong and Scaramuzza, Davide},
  journal={IEEE Transactions on Robotics},
  year={2022},
  publisher={IEEE}
}

@book{diaz20193d,
  title={3D Motion of Rigid Bodies},
  author={D{\'\i}az, Ernesto Olgu{\'\i}n and D{\'\i}az, Olgu{\'\i}n and Ditzinger},
  year={2019},
  publisher={Springer}
}

@article{deisenroth2013survey,
  title={A survey on policy search for robotics},
  author={Deisenroth, Marc Peter and Neumann, Gerhard and Peters, Jan and others},
  journal={Foundations and Trends{\textregistered} in Robotics},
  volume={2},
  number={1--2},
  pages={1--142},
  year={2013},
  publisher={Now Publishers, Inc.}
}

@article{andersson2019casadi,
  title={CasADi: a software framework for nonlinear optimization and optimal control},
  author={Andersson, Joel AE and Gillis, Joris and Horn, Greg and Rawlings, James B and Diehl, Moritz},
  journal={Mathematical Programming Computation},
  volume={11},
  number={1},
  pages={1--36},
  year={2019},
  publisher={Springer}
}

@article{paszke2019pytorch,
  title={Pytorch: An imperative style, high-performance deep learning library},
  author={Paszke, Adam and Gross, Sam and Massa, Francisco and Lerer, Adam and Bradbury, James and Chanan, Gregory and Killeen, Trevor and Lin, Zeming and Gimelshein, Natalia and Antiga, Luca and others},
  journal={Advances in neural information processing systems},
  volume={32},
  pages={8026--8037},
  year={2019}
}

@article{kingma2014adam,
  title={Adam: A method for stochastic optimization},
  author={Kingma, Diederik P and Ba, Jimmy},
  journal={arXiv preprint arXiv:1412.6980},
  year={2014}
}

\end{document}